\documentclass[runningheads]{llncs}
\usepackage{graphicx}

\usepackage{eccv}
\usepackage{tikz}
\usepackage{comment}
\usepackage{amsmath,amssymb} 
\usepackage{color}
\usepackage{multirow}
\usepackage{graphicx}
\usepackage{amsmath}
\usepackage{amssymb}
\usepackage{boldline}
\usepackage{algorithm}
\usepackage{algorithmic}

\usepackage{multicol}
\usepackage{booktabs}
\usepackage{float}

\usepackage[accsupp]{axessibility}  

\usepackage[width=122mm,left=12mm,paperwidth=146mm,height=193mm,top=12mm,paperheight=217mm]{geometry}
\usepackage{hyperref}
\usepackage{microtype}
 
\begin{document}
\pagestyle{headings}
\mainmatter

\title{REPS: Reconstruction-based Point Cloud Sampling} 

\titlerunning{ } 
\authorrunning{ } 
\author{
Guoqing Zhang, 
Wenbo Zhao, 
Jian Liu,
Xianming Liu 
}
\institute{Harbin Institute of Technology}

\maketitle

\begin{abstract} 
Sampling is widely used in various point cloud tasks as it can effectively reduce resource consumption. Recently, some methods have proposed utilizing neural networks to optimize the sampling process for various task requirements. Currently, deep downsampling methods can be categorized into two main types: generative-based and score-based. Generative-based methods directly generate sampled point clouds using networks, whereas score-based methods assess the importance of points according to specific rules and then select sampled point clouds based on their scores. However, these methods often result in noticeable clustering effects in high-intensity feature areas, compromising their ability to preserve small-scale features and leading to the loss of some structures, thereby affecting the performance of subsequent tasks. In this paper, we propose REPS, a reconstruction-based scoring strategy that evaluates the importance of each vertex by removing and reconstructing them using surrounding vertices. Our reconstruction process comprises point reconstruction and shape reconstruction. The two aforementioned reconstruction methods effectively evaluate the importance of vertices by removing them at different scales for reconstruction. These reconstructions ensure that our method maintains the overall geometric features of the point cloud and avoids disturbing small-scale structures during sampling. Additionally, we propose the Global-Local Fusion Attention (GLFA) module, which aggregates local and global attention features of point clouds, ensuring high-quality reconstruction and sampling effects. Our method outperforms previous approaches in preserving the structural features of the sampled point clouds. Furthermore, abundant experimental results demonstrate the superior performance of our method across various common tasks. Source code and weights are available at: \href{https://github.com/hitcslj/REPS}{https://github.com/hitcslj/REPS}.

\end{abstract}

\section{Introduction}
The advancement of 3D sensor technology has made acquiring point cloud data remarkably convenient. Point clouds are widely used in fields such as autonomous driving, robotics, and virtual reality. The substantial data volume of point clouds demands significant computational resources for processing. Hence, achieving efficient downsampling of point cloud data while ensuring performance becomes crucial.

Downsampling is a critical module in various point cloud models, playing a crucial role in improving model performance and reducing resource consumption. Currently, predominant methods for point cloud sampling primarily rely on conventional approaches, with random sampling and farthest point sampling \cite{eldar1997farthest} being representative solutions. Traditional methods offer flexibility as they can be integrated into various network models. However, their drawback is using fixed sampling patterns, which may generate point clouds incompatible with the requirements of different tasks. We believe that different tasks may have diverse requirements for sampled points. Traditional methods cannot dynamically adjust the sampling strategy according to the task's requirements. This limitation implicitly hinders the enhancement of model performance.

Some methods propose using neural networks to dynamically optimize downsampling for various tasks, replacing traditional approaches. Deep point cloud sampling methods can be broadly categorized as generation-based and scoring-based. Generation-based methods involve feeding the original point cloud into a designed sampling network, which then directly outputs the downsampled point cloud \cite{Dovrat_2019_CVPR, wang2021pst, lang2020samplenet, qian2020mops, lin2021net}. Scoring-based methods estimate the importance of each point through a certain mechanism, representing its importance, and then select sampling points directly from the original point cloud based on these scores according to certain rules \cite{hong2023attention, Wu_2023_CVPR}. However, both methods often produce significant clustering effects in regions with high feature intensity, leading to the inability to preserve smaller-scale features, resulting in feature loss and affecting the performance of subsequent tasks. The clustering effect causes the distribution of the sampled point cloud to become uneven. With a low sampling rate, these methods often fail to maintain the structure of small components, such as chair legs and car wheels.

Therefore, we propose REPS to address these issues by introducing a novel approach for assessing the importance of vertices. Through the implementation of point and shape reconstruction, we can accurately assess the importance of each point. In point reconstruction, the central point is reconstructed based on the neighborhood features of each point. Accurate reconstruction of a point indicates redundancy of the information it holds in the original point cloud. Moreover, points situated on the object's outline are typically more challenging to reconstruct compared to those in the central plane, leading to higher scores. Consequently, our point reconstruction method can delineate the object's outline and preserve its overall geometric characteristics. Furthermore, to ensure that points in flat regions of the object are adequately considered, we introduce shape reconstruction. Specifically, we initially choose a local region, randomly remove certain points from it, and reconstruct the entire shape using the remaining points. If accurate reconstruction of the complete shape is achievable after point removal, it suggests that the remaining points adequately represent the region's shape. As finer structures are often harder to reconstruct, points in these regions are assigned higher scores. This ensures the preservation of these structures during the sampling process. Furthermore, as the quality of downsampling depends on the reconstruction process, which in turn is determined by the point cloud features extracted by the model, we introduce the Global-Local Fusion Attention (GLFA) module. This module effectively integrates both local and global features of the point cloud, ensuring the quality of both reconstruction and downsampling.

In summary, the key contributions of this paper are threefold:
\begin{itemize}
\item We propose a novel reconstruction-based sampling scheme that effectively preserves the structural features of the downsampled point cloud.
\item We propose GLFA, which effectively aggregates local and global features, ensuring high-quality reconstruction and sampling effectiveness.
\item Extensive experiments demonstrate the performance and effectiveness of our REPS across various tasks and datasets.
\end{itemize}

\section{Related Work} 

\subsection{Point Cloud Sampling}
Downsampling is a crucial operation in managing large-scale point cloud data, effectively conserving computational resources. Presently, point cloud sampling solutions fall into traditional and learning-based categories.FPS (Farthest Point Sampling) is the predominant traditional method, selecting points iteratively based on their maximum distance from the already sampled points. This process gradually constructs a uniformly distributed set of sampled points, thereby maintaining the distribution of the original point cloud. However, when dealing with large-scale point cloud data, FPS is frequently substituted with the more efficient RS (Random Sampling). Nevertheless, both RS and FPS are task-agnostic, lacking the ability to dynamically adjust sampling strategies according to different task requirements. This limitation serves as a potential bottleneck, hindering further enhancement of model performance.

The key to achieving dynamic sampling lies in transforming the non-differentiable sampling process into a differentiable operation. Hence, learning-based methods \cite{Dovrat_2019_CVPR, wang2021pst, lang2020samplenet, qian2020mops, lin2021net, cheng2022meta, wang2023lightn, Wu_2023_CVPR, hong2023attention, wen2023learnable} have been proposed, with S-Net being one of the pioneering attempts. S-Net \cite{Dovrat_2019_CVPR} takes the original point cloud as input, processes it through several layers of simple MLP, and directly outputs a simplified point set. These points are then fed into the downstream task network for evaluation. Through backpropagation, the generated points are optimized for the given task. During inference, S-Net matches the network-generated points with the original point cloud to obtain the downsampled points. Building upon this, SampleNet \cite{lang2020samplenet} further optimized the process. Each point generated by the network is projected onto its k-nearest neighbors in the original point cloud. The resulting projected point cloud is then fed into the task network. \cite{wen2023learnable} propose a skeleton-aware point cloud sampling method to preserve the object geometry and topology information during sampling. LighTN \cite{wang2023lightn} attempts to incorporate attention mechanisms into point cloud sampling. The methods above are all generation-based, characterized by directly outputting sampled points from the network. Such approaches cannot be easily embedded flexibly into other models like traditional methods such as FPS. Furthermore, optimizing the sampled points necessitates training not only the task network but also an additional downsampling network, which may require additional computational resources. To address this, some methods \cite{Wu_2023_CVPR, hong2023attention} propose plug-and-play point cloud sampling modules. Inspired by edge detection in images, APES \cite{Wu_2023_CVPR} computes attention values within local neighborhoods of the point cloud and determines whether a point is an edge point based on the standard deviation of attention values, thus achieving the extraction of edge points from the point cloud. Score-based methods like APES \cite{Wu_2023_CVPR} can be flexibly integrated into various models and trained and optimized along with the model. Moreover, they often have fewer parameters and faster inference speeds. Our method effectively incorporates the supervision information of the original point cloud by implementing the above two reconstruction methods, which can also be seen as a form of local point cloud generation. Then, based on the reconstruction loss, we score each point and select sampling points, enabling our method to combine the flexibility of score-based methods. Therefore, we achieve a perfect integration of the above two sampling strategies.
\vspace{-0.6em}
\subsection{Deep Learning on Point Clouds}
Deep neural networks have been widely applied to tasks such as point cloud classification \cite{goyal2021revisiting, hamdi2021mvtn, liu2019point2sequence, ma2022rethinking, xiang2021walk}, segmentation \cite{behley2021towards, hu2020randla, landrieu2018large, hu2022sensaturban}, reconstruction \cite{fan2017point, mandikal20183d}, completion \cite{yuan2018pcn, yang2018foldingnet, xie2020grnet, zhou2022seedformer}, etc. Unlike structurally regular and easily manipulable 2D images, 3D point clouds represent an unordered set of points in space. The early migration of neural networks to the point cloud domain faced significant challenges. PointNet \cite{qi2017pointnet} innovatively utilizes symmetric functions to handle unordered point cloud data, achieving end-to-end point cloud learning. It represents the first successful integration of neural networks with point clouds. Building upon PointNet, PointNet++ \cite{qi2017pointnet++} introduced a hierarchical processing approach that more effectively captures both local and global features of point clouds, leading to improved performance in classification and segmentation tasks. Subsequently, the gateway to deep learning for point clouds was opened, and an increasing number of endeavors began to apply the successful experiences of neural networks in images to point clouds. In the exploration of efficiently extracting local structural features from point clouds, some directions have emerged. Among them, KPConv \cite{thomas2019kpconv} introduced a novel convolutional operation that effectively captures local structural information in point clouds, exhibiting strong rotation and scale invariance. DGCNN \cite{wang2019dynamic} significantly enhances the model's expressive power by implementing a dynamic graph convolutional neural network. Recently, some works \cite{engel2021point, zhao2021point, guo2021pct, wu2023point} have started combining attention mechanisms with point clouds. By optimizing point cloud structures through spatial filling curves, PTv3 \cite{wu2023point} significantly improves the performance of Transformers in point cloud processing, achieving larger, faster, and more powerful point cloud models.

\section{Method} 
Given an input point cloud with $N$ points ~$\mathcal{P}=\{\mathbf{p}_i\}_{i=1}^N$ and corresponding point features ~$\mathcal{X}=\{\mathbf{x}_i\}_{i=1}^N$, where $\mathbf{p}_i\in\mathbb{R}^3$ represents the spatial coordinates of the $i$-th point, and $\mathbf{x}_i\in\mathbb{R}^d$ corresponds to its features. Our goal is to obtain a simplified point set with $M$ ($M < N$) points ~$\mathcal{Q}=\{\mathbf{q}_i\}_{i=1}^M$, where $r=\frac{N}{M}$ represents the downsampling rate. In this section, we will provide a detailed description of our REPS. First, we explain the implementation of point reconstruction (Sec. \ref{Sec3.1}) and shape reconstruction (Sec. \ref{sec3.2}). Following these reconstructions, we will provide a detailed description of the downsampling process (Sec. \ref{Sec3.3}). Next, we illustrate details of our point cloud feature extractor, GLFA module(Sec. \ref{Sec3.4}). Finally, we present the overall architecture of our method (Sec. \ref{Sec3.5}).

\begin{figure}[t]
	\begin{center}
		\includegraphics[width=0.9\linewidth]{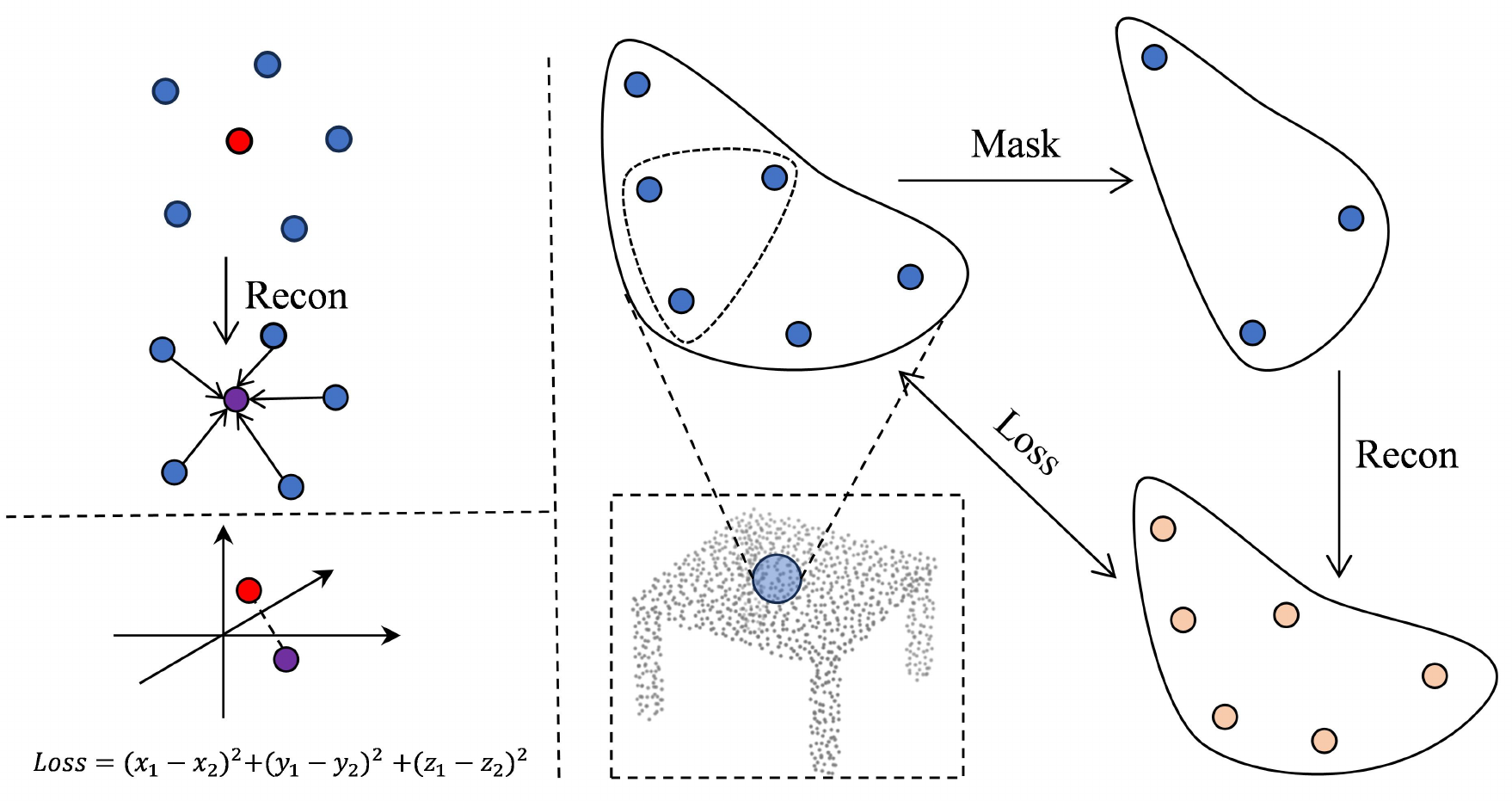}
	\end{center}
	\vspace{-0.3cm}
    \caption{\textbf{Left: Point Reconstruction.} For each point in the input point cloud, we select its K nearest neighbors and then reconstruct the central point based on the features of these neighboring points. \textbf{Right: Shape Reconstruction.} For each local patch, we randomly remove some points and then use the remaining points to reconstruct the complete local shape.}
	\label{fig: recon}
\vspace{-0.5cm}
\end{figure}

\vspace{-0.5em}
\subsection{Point Reconstruction} \label{Sec3.1} 
We first introduce point reconstruction to assess the importance of vertices, which effectively extract the contour features of objects. For each $ \mathbf{p}_i \in \mathcal{P} $, we attempt to reconstruct it. As shown in the Figure \ref{fig: recon}, for $ p_i $, we select its $K$ nearest neighbors:
\begin{equation} 
    \mathcal{N}_i = k\text{NN}(\mathcal{P},\mathbf{p}_i), 
    \label{eq:knn}
\end{equation}
where \( \mathcal{N}_i \) represents the indices of its K nearest neighbors. Then, we respectively retrieve the positions and features of the $K$ nearest neighbors:
\begin{equation}
    \begin{aligned}
         \mathcal{P}_i &= \text{Select}(\mathcal{P},\mathcal{N}_i)\\
         \mathcal{F}_i &= \text{Select}(\mathcal{X},\mathcal{N}_i)
    \end{aligned}
\end{equation}  
Finally, based on the information mentioned above, we reconstruct the spatial coordinates of $\mathbf{p}_i$ to obtain the reconstructed point set $\hat{\mathcal{P}}$:
\begin{equation} 
    \hat{\mathcal{P}} = \{\hat{\mathbf{p}_i} \mid \hat{\mathbf{p}_i} = \operatorname{MLP}(\mathcal{P}_i, \mathcal{F}_i) \}_{i=1}^N
    \label{eq:recp}
\end{equation}
If a point can be precisely reconstructed using its $K$ nearest neighbors, the information in that point is redundant, and removing it has a negligible impact.
\vspace{-1.5em}
\subsection{Shape Reconstruction} \label{sec3.2}
\vspace{-0.75em}
Since point reconstruction focuses more on contour points, to ensure that points in flat regions also receive attention, we propose shape reconstruction. To perform shape reconstruction, we first partition the point cloud into $M$ local patches $\mathcal{H}=\{H_j\}_{j=1}^M$, each with a size of $K$. Next, we randomly divide $H_j$ into two subsets, representing the points to be removed and the points to be retained, respectively:
\begin{equation}
    \begin{aligned} 
      &  {H}_j^{remove} = \{\mathbf{q}_i \mid \mathbf{q}_i \in H_j \}_{i=1}^{K/2} \\
      &  {H}_j^{retain} = \{\mathbf{q}_i \mid \mathbf{q}_i \in H_j \setminus {H}_j^{remove} \}_{i=1}^{K/2}
    \end{aligned} 
\end{equation}
Our goal is to reconstruct the complete local shape using the remaining points. Similarly, we first retrieve the positions and features of the retained points:
\begin{equation}
    \begin{aligned}
         \mathcal{P}'_j &= \{p'_i = \text{Select}(\mathcal{P}, \text{Index}(\mathbf{q}_i)) \mid \mathbf{q}_i \in H_j^{retain} \} \\
         \mathcal{F}'_j &= \{f'_i = \text{Select}(\mathcal{X}, \text{Index}(\mathbf{q}_i)) \mid \mathbf{q}_i \in H_j^{retain} \}
    \end{aligned}
\end{equation} 
Then, based on the features of the remaining points, we reconstruct the complete local patch. It can be formalized as follows:  
\begin{equation}
\begin{aligned}
    \hat{H}_j = \{\hat{\mathbf{q}}_i \mid \hat{\mathbf{q}}_i = \operatorname{MLP}(\mathcal{P}'_j, \mathcal{F}'_j) \}_{i=1}^{K} \\
    \hat{\mathcal{H}} = \{\hat{H}_1,\hat{H}_2,...,\hat{H}_j,...\} \quad\quad
\end{aligned} 
\end{equation}  
In our perspective, if it is difficult to reconstruct the local shape after removing some points, it indicates that these points are crucial for maintaining the shape. On the other hand, if the local shape can be well reconstructed with the remaining points, it suggests that the removal of points has a minimal impact, and the remaining points have effectively preserved the overall shape information. 
 
\vspace{-2.0em}
\subsection{Reconstruction-based Sampling} \label{Sec3.3} 
\vspace{-1.0em}
After completing the aforementioned reconstruction, we obtain the losses for the reconstructed point and the shape, respectively:
\begin{equation}
    \begin{aligned}
        \mathcal{L}oss_{point} &= \{ \| \mathbf{p}_i - \mathbf{\hat{p}_i} \|_2 \mid \mathbf{p}_i \in \mathcal{P}, \mathbf{\hat{p}_i} \in \mathcal{\hat{P}} \}_{i=1}^{N} \\
        \mathcal{L}oss_{shape} &= \{ \sum_{k=1}^{K} \| \mathbf{q}_k - \mathbf{\hat{q}}_k) \|_2 \mid \mathbf{q}_k \in \mathcal{H}_j, \mathbf{\hat{q}}_k \in \mathcal{\hat{H}}_j \}_{j=1}^{M}
    \end{aligned}
\end{equation}
Afterward, we score each point based on these losses, and the specific process is outlined in Algorithm \ref{alg:alg1}.
\begin{center}  
\begin{minipage}{.8\linewidth}  
\vspace{-2.0em}
\begin{algorithm}[H]  
    \caption{Reconstruction-based Scoring}
    \label{alg:alg1} 
    \begin{algorithmic}[1] 
        \STATE \textbf{INPUT}:$\mathcal{L}oss_{point}, \mathcal{L}oss_{shape}, \mathcal{P}, \mathcal{H}$
        \STATE N$\leftarrow{\mid\mathcal{P}\mid}$, M$\leftarrow{\mid\mathcal{H}\mid}$
        \STATE $Score_{point}\leftarrow$ zeros(N), $Score_{shape}\leftarrow$ zeros(N)
        \STATE $Score_{point}\leftarrow \mathcal{L}oss_{point}$
        \FOR{$j\leftarrow1$ \TO M} 
            \FOR{$\mathbf{q}$ in ${H}_j^{remove}$}  
	        \STATE $id$ = \text{Index}($\mathbf{q}$) 
                \STATE $Score_{shape}[id]\leftarrow$ $\mathcal{L}oss_{shape}[j]$ 
            \ENDFOR
            \FOR{$\mathbf{q}$ in ${H}_j^{retain}$} 
	        \STATE $id$ = \text{Index}($\mathbf{q}$) 
                \STATE $Score_{shape}[id]\leftarrow$ 1 - $\mathcal{L}oss_{shape}[j]$ 
            \ENDFOR 
        \ENDFOR
        \STATE $Score_{total} \leftarrow \alpha Score_{point} + (1-\alpha) Score_{shape}$
        \STATE \textbf{OUTPUT:} $Score_{total}$
    \end{algorithmic} 
\end{algorithm} 
\end{minipage} 
\end{center} 

\begin{figure}[h]
\vspace{-2.0em}
	\begin{center}
		\includegraphics[width=0.95\linewidth, height=0.5\linewidth]{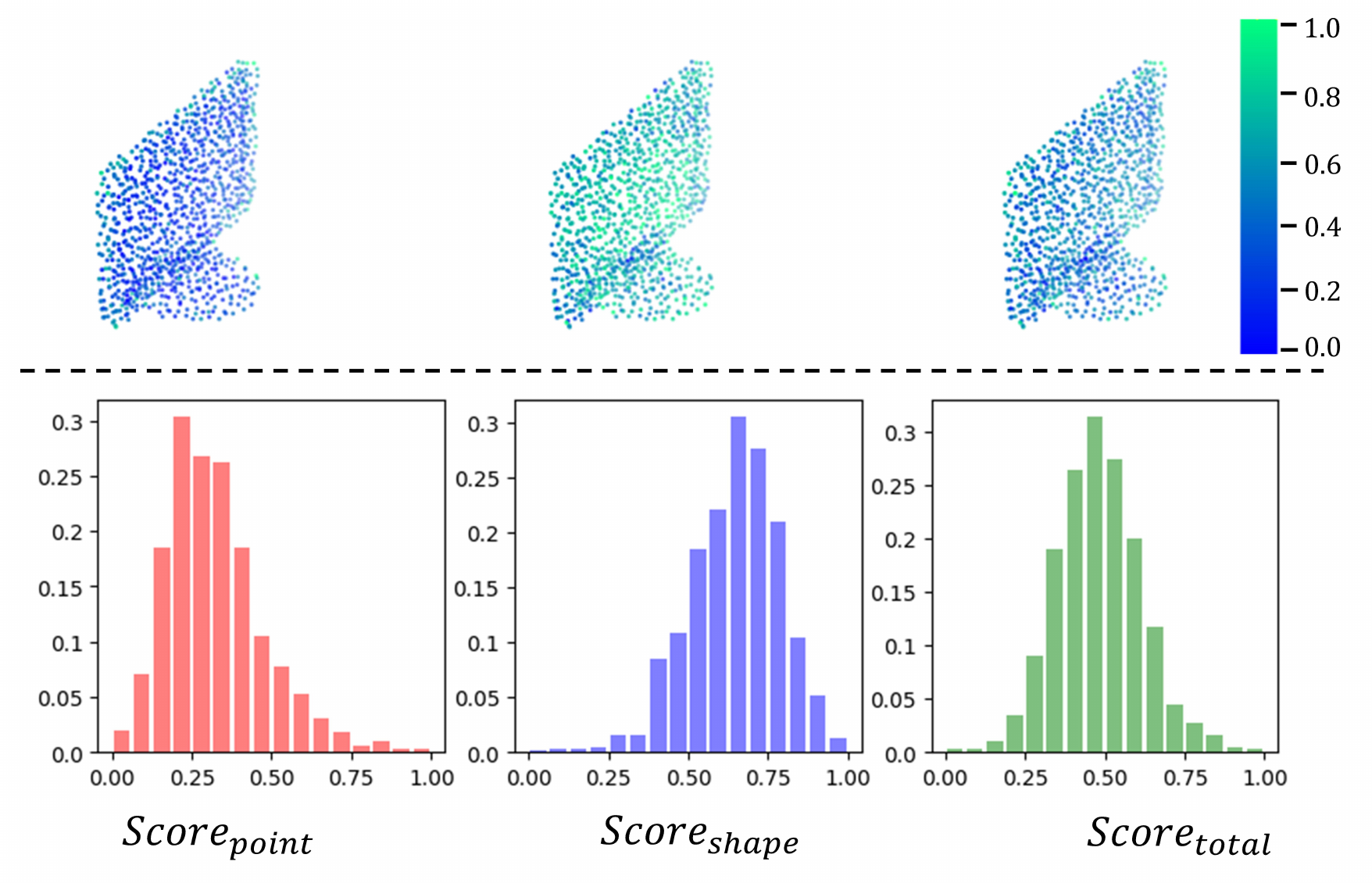}
	\end{center}
	\vspace{-0.3cm}
    \caption{Top: Visualization of scores based on reconstruction. Bottom: Distribution of scores based on point reconstruction, scores based on shape reconstruction, and total scores.}
	\label{fig: score}
\vspace{-2.0em}
\end{figure}

Through the above process, we have achieved the goal of assigning a score to each point, and the results are illustrated in Figure \ref{fig: score}. From the figure, it can be observed that the points emphasized by the reconstruction based on points and the reconstruction based on shape have distinct differences. The reconstruction based on points tends to focus more on edge points, while the shape-based reconstruction pays more attention to preserving local shapes such as planes. 

Now, with scores assigned to each point representing its importance in the point cloud, to complete the downsampling operation, we naturally choose the points with the highest scores as the final sampled points:
\begin{equation}
    \begin{aligned}
        \text{indices} &= \text{argmax}(\mathcal{S}core_{total}, \text{M})\\
        \mathcal{Q} &= \text{Select}(\mathcal{P}, \text{indices})
    \end{aligned}
\end{equation}
These points not only preserve the geometric information of the original point cloud but also effectively retain the features learned by the network.
\subsection{GLFA Module} \label{Sec3.4}
Both of the aforementioned strategies are built upon reconstruction, and the quality of the reconstruction results depends crucially on the features extracted by the feature extraction module. Therefore, we propose GLFA (Global-Local Fusion Attention), which can effectively integrate local and global structural features, enabling effective reconstruction operations and thereby ensuring the quality of the sampled point cloud. 

For the input point cloud $\mathcal{P}$, after performing the downsampling operation, we obtain the downsampled point cloud $\mathcal{Q}$. To extract local features, first, for each $\mathbf{q}_i$ in $\mathcal{Q}$, we query its K nearest neighbors in $\mathcal{P}$:
\begin{equation} 
    \begin{aligned} 
        \mathcal{L}_i &= \text{Select}(\mathcal{X}, k\text{NN}(\mathcal{P},\mathbf{q}_i))
    \end{aligned} 
    \label{eq:knn2}
\end{equation}
where $\mathcal{L}_i \in R^{K \times d}$ represents the features corresponding to the K nearest neighbors, they describe the shape features around the sampled points. Next, for each $\mathcal{L}_i$, we perform self-attention operations to extract local structural features.:
\begin{equation} 
    \mathcal{L}_i = \text{self-attn}(\mathcal{L}_i) \in R^{K \times d} 
\end{equation}
where $\text{self-attn}(\mathcal{\text{X}}) = \text{softmax}(\frac{\text{X}\text{X}^T} {\sqrt{d}})\text{X}$ represents the attention operation. Then, to aggregate local features, we perform a pooling operation for each $\mathcal{L}_i$. Furthermore, to achieve information exchange among different local shapes, we perform a global attention operation on all $\mathcal{L}_i$.:
\begin{equation} 
\begin{aligned}
    \mathcal{L} &= \{\text{pooling}(\mathcal{L}_i) \}_{i=1}^{M} \in R^{M \times d} \\
    \mathcal{L} &=  \text{self-attn}(\mathcal{L}) \in R^{M \times d} 
\end{aligned}
\end{equation}
Through the above operations, our GLFA module can effectively incorporate both local and global features of objects.
\subsection{Loss Function} \label{Sec3.5}
Our loss function consists of two components: the sampling loss and the task loss. The sampling loss is defined as follows:
\begin{equation}
    \mathcal{L}oss_{sample} = \sum_{i=1}^{N}{\mathcal{L}oss_{point}^i} + \sum_{i=1}^{M}{\mathcal{L}oss_{shape}^i}
\end{equation}
The ultimate total loss is formulated as follows:
\begin{equation}
    \mathcal{L}oss_{total} = \mathcal{L}oss_{sample} + \mathcal{L}oss_{task}
\end{equation}
where $\mathcal{L}oss_{task}$ is the loss corresponding to the downstream task (e.g., classification).

\begin{figure}[t]
	\begin{center}
		\includegraphics[width=0.99\linewidth]{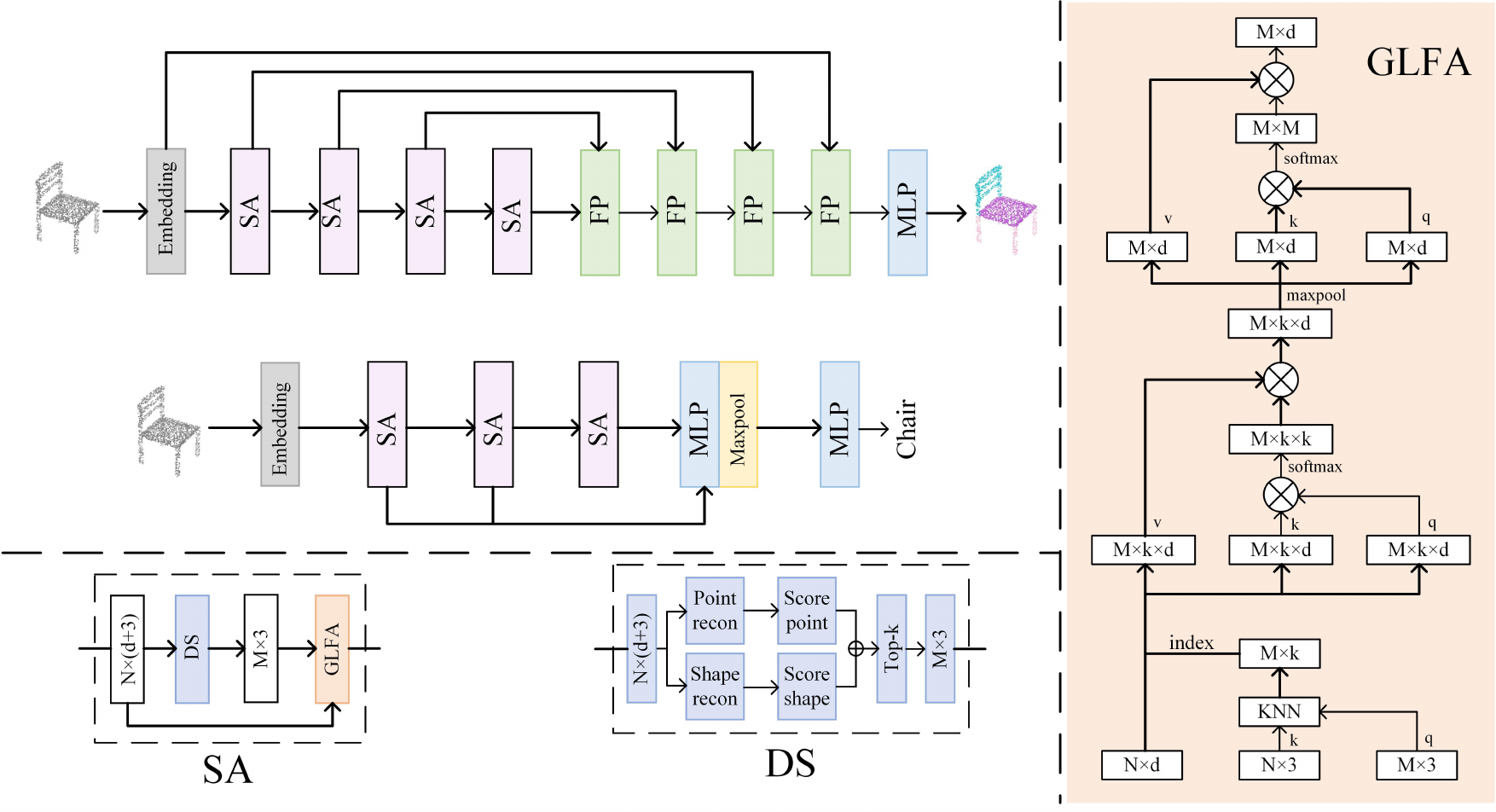}
	\end{center}
	\vspace{-0.3cm}
    \caption{The overview of our method. Left: Architecture of semantic segmentation and classification models. Right: Implementation details of the GLFA module.}
	\label{fig: over}
\vspace{-2.0em}
\end{figure}
\begin{figure*}[ht]
	\begin{center}
		\includegraphics[width=0.9\linewidth]{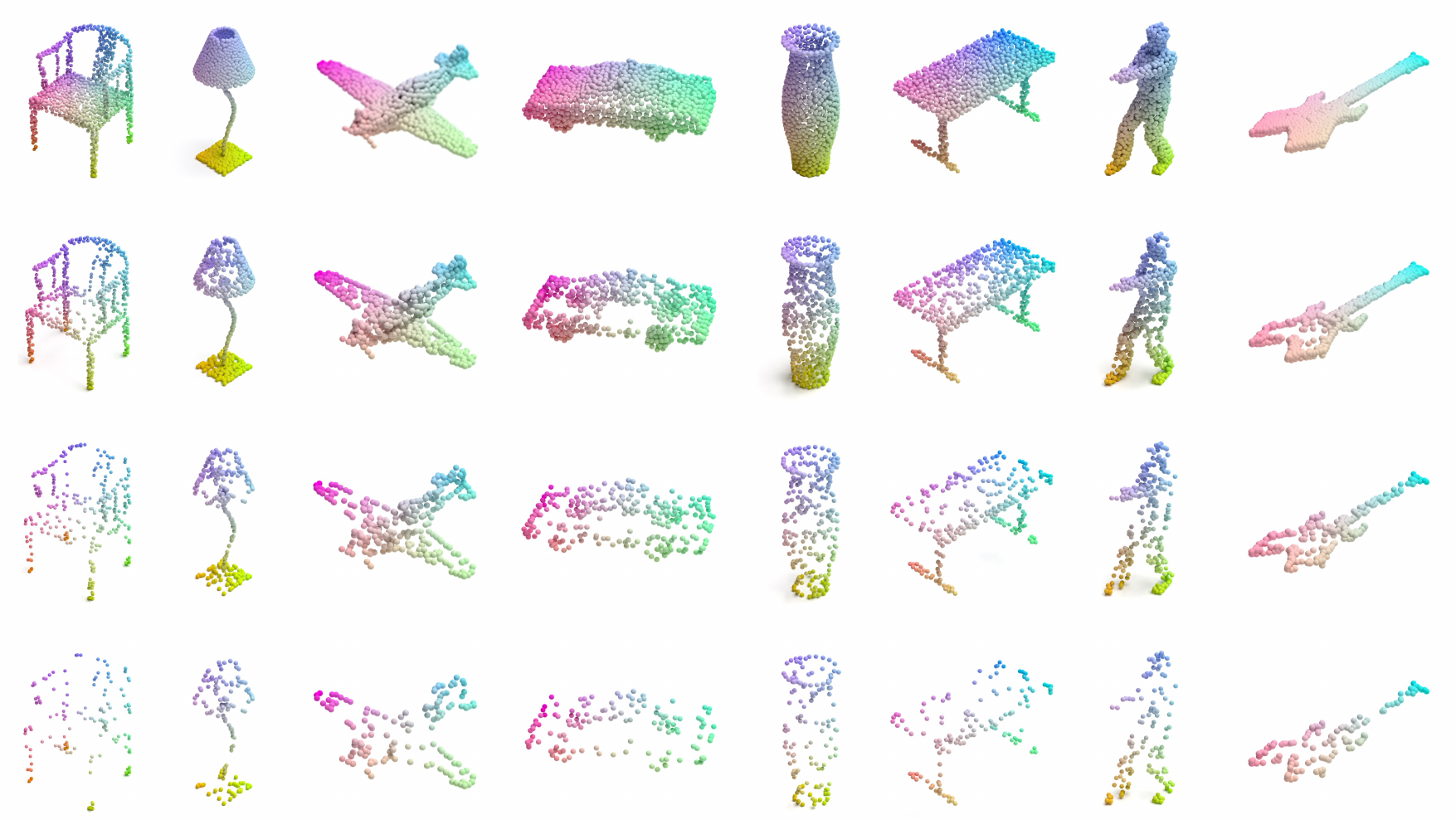}
	\end{center}
	\vspace{-0.3cm}
    \caption{\textbf{Point cloud sampling visualization results on the ModelNet40 dataset.} At the top is the original point cloud (1024 points), followed by downsampled point clouds with sampling ratios of 0.5, 0.25, and 0.125, respectively.}
	\label{fig:vis1} 
\vspace{-2.0em}
\end{figure*}

\section{Experiments} 
\subsection{Shape Classification}

\smallskip\noindent\textbf{Dataset.} ModelNet40 \cite{wu20153d} is currently a widely used benchmark dataset for evaluating point cloud classification. It consists of 12,311 object point clouds, uniformly sampled from the surface of 3D meshes and classified into 40 categories. For fairness, we follow the official train-test split, where 9,843 point clouds are used for training and 2,468 point clouds are used for testing. Moreover, only 3D coordinates serve as the original input. Regarding data augmentation, we follow the settings of APES \cite{Wu_2023_CVPR}. During testing, we do not employ any data augmentation techniques or voting methods.

\smallskip\noindent\textbf{Network Architecture.} The architecture of our classification network is illustrated in Figure \ref{fig: over}. First, the input point cloud is mapped to a high-dimensional space through the embedding layer. Then, the Set Abstraction module, composed of downsampling and GLFA, is utilized to further extract point cloud geometric features. Moreover, for an input point cloud of size N, all sampling modules downsample it to N/2 points. Finally, the max-pooling operation is used to obtain global features, which are then fed into an MLP to obtain the final prediction. In particular, we set the neighborhood size \(K\) to 16, and \(\alpha\) to 0.8.
 
\smallskip\noindent\textbf{Quantitative and Qualitative Results.}
The quantitative results for point cloud classification are presented in Table \ref{table:cls_seg}. It can be observed that our proposed REPS achieves SOTA performance compared to other classical classification methods. Moreover, in comparison to the similar method APES \cite{Wu_2023_CVPR}, we also demonstrate significant advantages. Figure \ref{fig:vis1} illustrates the Qualitative results of our method on point cloud downsampling. All downsampled point clouds exhibit visually pleasing results. Even with an increase in the downsampling factor, we can still intuitively discern the geometric features of the original object on the downsampled point clouds. Furthermore, our method performs well in preserving object details. For instance, small components like car wheels and chair legs, which occupy a relatively small proportion in the original point cloud, are still largely preserved in the downsampled point clouds generated by our method.

\begin{table}[t]
\centering
\resizebox{0.85\columnwidth}{!}{
\setlength\tabcolsep{2pt} 
\begin{tabular}{ccc}
\toprule
Method & \quad points\quad\quad & OA \\ \midrule
PointNet \cite{qi2017pointnet} & 1024 &89.2\% \\
PointNet++ \cite{qi2017pointnet++} & 1024 & 91.9\% \\
DGCNN \cite{wang2019dynamic} & 1024 &92.9\% \\
PointCNN \cite{li2018pointcnn} & 1024 &92.2\% \\
PointConv \cite{wu2019pointconv} & 1024 &92.5\% \\
PVCNN \cite{liu2019point} & 1024 &92.4\% \\
KPConv \cite{thomas2019kpconv} & 1024 &92.9\% \\
PointASNL \cite{yan2020pointasnl} & 1024 &93.2\% \\ 
PT \cite{zhao2021point} & 1024 &93.7\% \\
PCT \cite{guo2021pct} & 1024 &93.2\% \\
PRA-Net \cite{cheng2021net} & 1024 &93.7\% \\
PAConv \cite{xu2021paconv}  & 1024 &93.6\% \\
CurveNet \cite{xiang2021walk} & 1024 &93.8\% \\
PointMLP \cite{ma2022rethinking} & 1024 &\textbf{94.1\%} \\ \midrule
APES (global) \cite{Wu_2023_CVPR}  & 2048 &93.8\% \\ 
REPS(ours) & 1024 & \textbf{94.1\%} \\ \bottomrule
\end{tabular}
\quad\quad
\begin{tabular}{lcc}
\toprule
Method & points & mIoU \\ \midrule
PointNet \cite{qi2017pointnet} & 2048 & 83.7\% \\
PointNet++ \cite{qi2017pointnet++} & 2048 & 85.1\% \\
SpiderCNN \cite{xu2018spidercnn} & 2048 & 85.3\% \\
DGCNN \cite{wang2019dynamic} & 2048 & 85.2\% \\
SPLATNet \cite{su2018splatnet} & 2048 & 85.4\% \\
PointCNN \cite{li2018pointcnn} & 2048 & 86.1\% \\
PointConv \cite{wu2019pointconv} & 2048 & 85.7\% \\
KPConv \cite{thomas2019kpconv} & 2048 & 86.2\% \\
PT \cite{zhao2021point} & 2048 & 86.6\% \\
PCT \cite{guo2021pct} & 2048 & 86.4\% \\
PRA-Net \cite{cheng2021net} & 2048 & 86.3\% \\
PAConv \cite{xu2021paconv} & 2048 & 86.1\% \\
CurveNet \cite{xiang2021walk} & 2048 & 86.6\% \\
StratifiedTransformer \cite{lai2022stratified} & 2048 & 86.6\% \\
\midrule
APES (global) \cite{Wu_2023_CVPR} & 2048 & 85.8\% \\
REPS(ours) & 2048 & \textbf{86.7\%}  \\ \bottomrule
\end{tabular}
} 
\caption{Left: Classification results on ModelNet40. Right: Segmentation results on the ShapeNetPart dataset. Note that our reported results did not consider the voting strategy.}
\label{table:cls_seg}
\vspace{-3.0em}
\end{table}

\subsection{Part Segmentation}
\smallskip\noindent\textbf{Dataset.} We evaluated the segmentation performance of our model on the ShapeNetPart \cite{yi2016scalable} dataset, which comprises 16 categories with a total of 16,881 models. Each category is labeled with 2 to 6 parts, resulting in a total of 50 parts. Each point cloud consists of 2048 points. We used 14,006 point clouds for training and 2,874 point clouds for testing. The mean intersection of union (mIoU) is used as a metric to evaluate the segmentation results.  

\begin{figure}[ht]
	\begin{center}
		\includegraphics[width=0.95\linewidth]{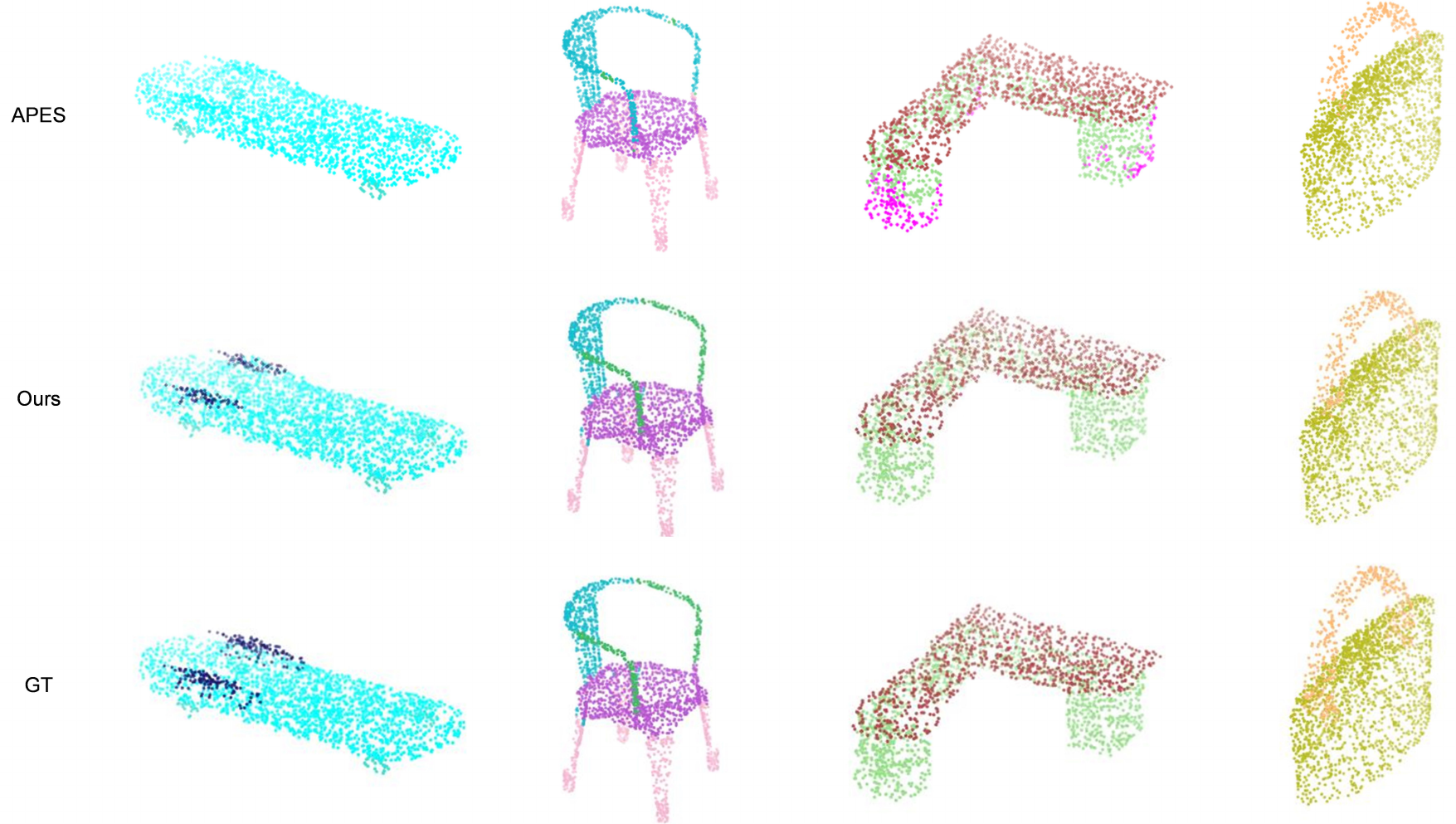}
	\end{center}
	\vspace{-0.3cm}
    \caption{Visualizations of segmentation results on the ShapeNetPart dataset.}
	\label{fig:seg} 
\vspace{-0.5cm}
\end{figure}
\smallskip\noindent\textbf{Network Architecture.} Following PointNet++ \cite{qi2017pointnet++}, the design of our segmentation network adopts the U-Net architecture. The network is structured with an encoder and a decoder. The encoder, composed of stacked sampling and GLFA modules, extracts multi-scale features. On the decoding side, the feature propagation module facilitates transmitting features from layer L to layer L-1, achieving the fusion of features at different scales. Ultimately, all point features are fed into the segmentation head for semantic label recognition. Specifically, we set K to 32 and $\alpha$ to 0.6.

\smallskip\noindent\textbf{Quantitative and Qualitative Results.} Table \ref{table:cls_seg} illustrates the results of our method on the point cloud semantic segmentation task. Compared to other classical deep point cloud segmentation methods, our REPS also achieves state-of-the-art performance. The advantage of our method over APES \cite{Wu_2023_CVPR} is evident. Figure \ref{fig:seg} illustrates the semantic segmentation results of our method on the ShapeNetPart dataset. Compared to APES \cite{Wu_2023_CVPR}, our segmentation results are more robust and accurate in some details.

\begin{table*}[b]
\vspace{-1.0em}
\centering
\resizebox{1\linewidth}{!}{
\begin{tabular}{c|ccccccccccc}
\toprule
$M$ & Voxel & RS & FPS \cite{eldar1997farthest} & S-NET \cite{Dovrat_2019_CVPR} & PST-NET \cite{wang2021pst} & SampleNet \cite{lang2020samplenet} & MOPS-Net \cite{qian2020mops} & DA-Net \cite{lin2021net} & LighTN \cite{wang2023lightn} & APES (global) \cite{Wu_2023_CVPR}  & REPS \\ \midrule
512 & 73.82 & 87.52 & 88.34 & 87.80 & 87.94 & 88.16 & 86.67 & 89.01 & 89.91 & 90.81 & \textbf{90.95}\\
256 & 73.50 & 77.09 & 83.64 & 82.38 & 83.15 & 84.27 & 86.63 & 86.24 & 88.21 & 90.40 & \textbf{90.79} \\
128 & 68.15 & 56.44 & 70.34 & 77.53 & 80.11 & 80.75 & 86.06 & 85.67 & 86.26 & 89.77 & \textbf{90.18} \\
64 & 58.31 & 31.69 & 46.42 & 70.45 & 76.06 & 79.86 & 85.25 & 85.55 & 86.51 & 89.57 & \textbf{89.89} \\
32 & 20.02 & 16.35 & 26.58 & 60.70 & 63.92 & 77.31 & 84.28 & 85.11 & 86.18 & 88.56 & \textbf{88.76} \\ 
\bottomrule
\end{tabular}}
\caption{Comparison with other sampling methods. Evaluated on the ModelNet40 classification benchmark with multiple sampling sizes. 
\vspace{0.5cm}
}
\label{table:compare} 
\vspace{-1.5cm}
\end{table*} 
\begin{figure*}[h]
	\begin{center}
		\includegraphics[width=0.95\linewidth]{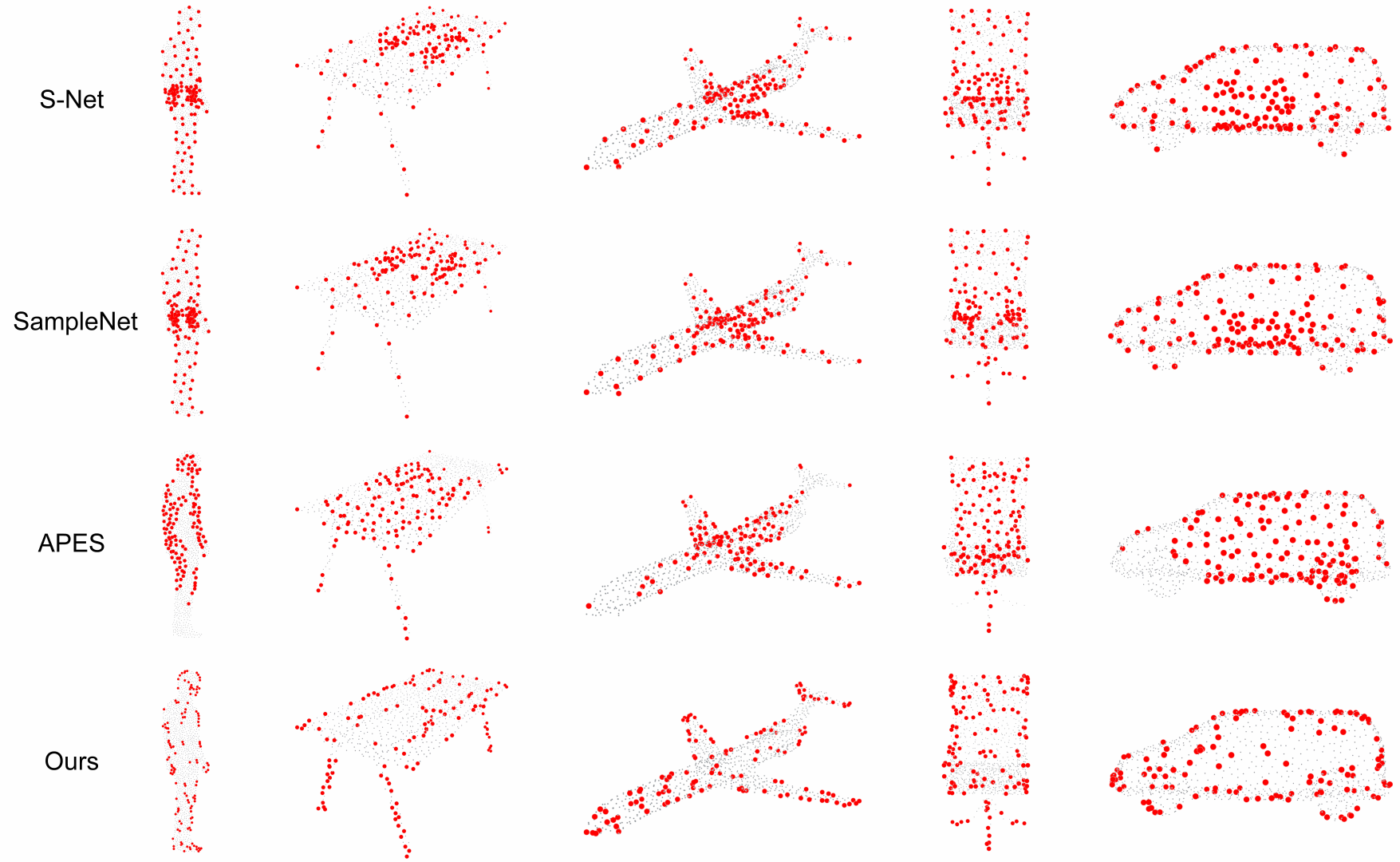}
	\end{center}
	\vspace{-0.3cm}
    \caption{Qualitative comparison for the sampling of 128 points from input point clouds with 1024 points.}
	\label{fig: vis2}  
\end{figure*}
\subsection{Sampling Comparison}
\smallskip\noindent\textbf{Experiment Setting.} To validate the effectiveness of our sampling method, we conducted a comprehensive comparison with previous works, including traditional RS, FPS, and learning-based methods such as S-Net \cite{Dovrat_2019_CVPR}, APES \cite{Wu_2023_CVPR}, etc. We adopted the same evaluation framework as APES \cite{Wu_2023_CVPR}: points sampled by our method are directly fed into downstream networks for evaluation. Specifically, we used the ModelNet40 classification task, with the PointNet as the task network. We evaluated the performance of different methods under various downsampling rates.

As shown in Figure \ref{fig:vis1}, the distribution of the downsampled point clouds undergoes significant changes with the decrease in the downsampling rate. Therefore, for a fair comparison, we adopt the same sampling strategy as APES \cite{Wu_2023_CVPR}. That is, to downsample an original point cloud of size N to M points, we first use FPS to sample it to 2M points and then apply our method to downsample it to M points.

\smallskip\noindent\textbf{Quantitative and Qualitative Results.} Table \ref{table:compare} presents the performance of each method on the task network under different downsampling rates. It can be observed that our method achieves state-of-the-art results at various downsampling rates. Figure \ref{fig: vis2} illustrates the downsampling results of different methods. Compared to other methods, our approach more intuitively showcases the geometric features of the original point cloud. Moreover, from the figure, it can be observed that methods like S-Net \cite{Dovrat_2019_CVPR} and SampleNet \cite{lang2020samplenet}, which are based on generation, exhibit a phenomenon of abnormal aggregation with many points concentrated in a specific local region, resulting in an uneven distribution of sampled points. APES \cite{Wu_2023_CVPR} also shows structural loss. Our method effectively addresses these issues. From the figure, it can be seen that some fine structures, such as car wheels and chair legs, are preserved in terms of geometric structure after being sampled by our method, which other methods fail to achieve.

\subsection{Ablation study}

\smallskip\noindent\textbf{Feature Learning Layer.} To further validate the effectiveness of our feature extraction module GLFA, as shown in Figure \ref{fig: over}, we conducted comparative experiments using different feature extraction methods. The results are presented in Table \ref{table:layers}, where it can be observed that our GLFA outperforms other methods. Additionally, when using EdgeConv as the feature extraction layer, our approach achieves performance improvement compared to the original DGCNN \cite{wang2019dynamic}.  

\begin{table}[t] 
\centering
\resizebox{0.99\columnwidth}{!}{%
\begin{tabular}{cccc}
\hline
\multirow{2}{*}{Method} & \multirow{2}{*}{Feature Learning Layer} & \multicolumn{2}{c}{points}        \\ \cline{3-4} 
                        &                                         & 1k              & 2k              \\ \hline
DGCNN                   & EdgeConv                               & 92.9\%          & 93.5\%          \\ \hline
\multirow{3}{*}{REPS}   & EdgeConv                                 & 93.2\%          & 93.7\%          \\
                        & N2P (APES)                           & 93.5\%          & 93.7\%          \\
                        & GLFA (Ours)          & \textbf{94.1\%} & \textbf{94.3\%} \\ \hline
\end{tabular}%
\quad\quad

\begin{tabular}{cccc}
\hline
\multicolumn{4}{l}{\quad Reconstruction}        \\ \cline{1-2}
\quad Point \quad & \quad Shape \quad & Cls.OA(\%) & Seg.mIoU(\%) \\ \hline
  \checkmark    &       &     93.91      &        86.37      \\ \hline
      &      \checkmark  &    93.22      &        85.90     \\ \hline
  \checkmark    &  \checkmark     &   \textbf{94.12}          &     \textbf{86.65}         \\ \hline
\end{tabular}%

}
\caption{Left: Ablation study of using different feature learning layers in the classification network. Right: Ablation study on point and shape reconstruction}
\label{table:layers}
\vspace{-3.2em}
\end{table}  

\begin{table}[h]
\vspace{-0.5em}
\centering
\resizebox{0.6\columnwidth}{!}{%
\begin{tabular}{cccccc}
\hline
K  & 4       & 8       & 16               & 32      & 64      \\ \hline
OA & 93.26\% & 93.43\% & \textbf{94.12\%} & 93.71\% & 93.83\% \\ \hline
\end{tabular}%
}
\caption{Ablation study of using a different number of neighbors.} 
\label{table:numk}
\vspace{-1.5em}
\end{table}
\smallskip\noindent\textbf{The effectiveness of point reconstruction and shape reconstruction.} We conducted ablation experiments on them to validate the effectiveness of using point reconstruction and shape reconstruction for sampling point selection. The results are shown in the table, indicating that adopting either strategy alone cannot achieve optimal performance. Additionally, it can be observed from the results that point-based reconstruction has a greater impact on model performance, perhaps since point-based reconstruction sampling focuses more on the object's contour features, ensuring the basic geometric structure of the object. Meanwhile, shape reconstruction allows points neglected by point reconstruction to have the opportunity to be sampled, increasing the diversity of the sampled point cloud and further improving performance.

\smallskip\noindent\textbf{The size of the K nearest neighbor (KNN) neighborhood.} Whether performing sampling operations or local feature extraction, the neighborhood size K plays an important role, so we compared the impact of K values of different sizes on performance, and the results are shown in Table \ref{table:numk}.
 
\begin{figure}[h]  
	\begin{center}
		\includegraphics[width=0.95\linewidth]{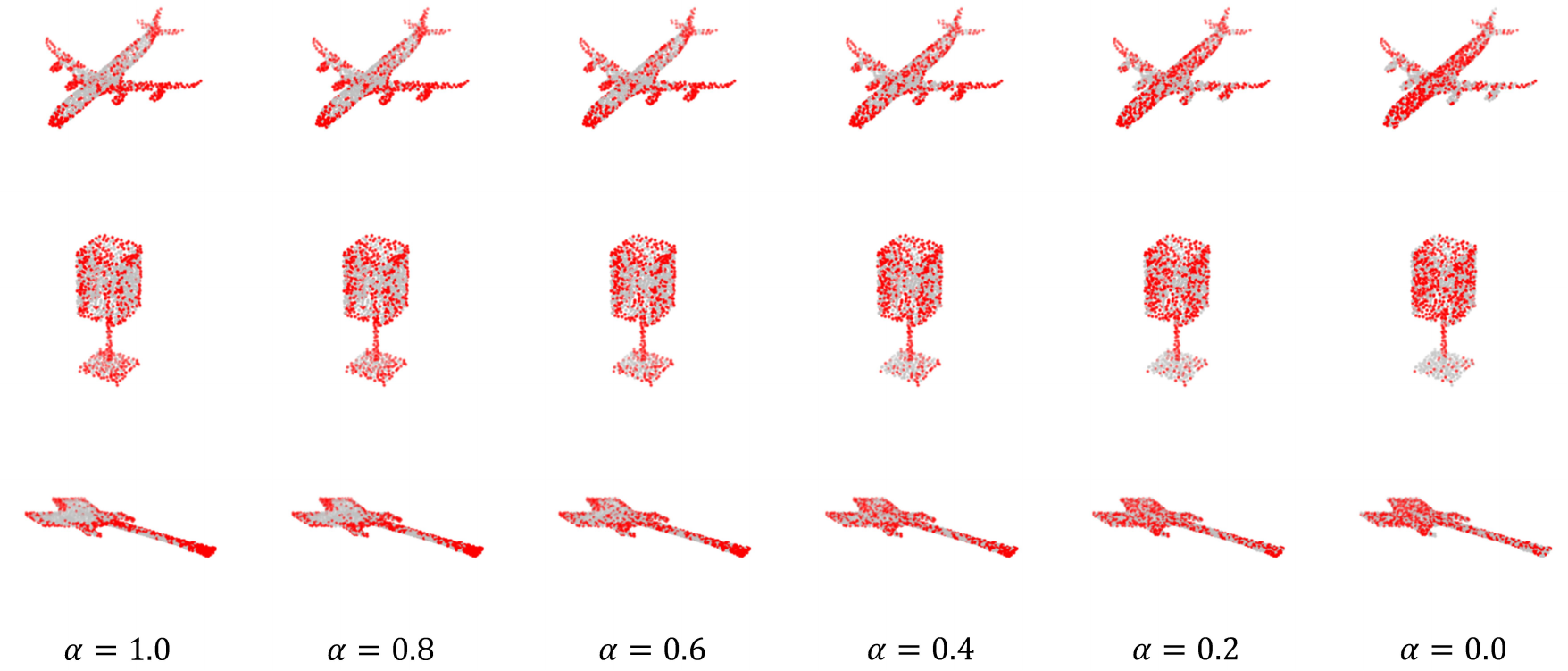}
	\end{center}
	\vspace{-0.3cm}
    \caption{Visualizing results of downsampling with different weights applied.}
	\label{fig: vis3}
\vspace{-2.0em}
\end{figure}
\smallskip\noindent\textbf{The impact of score weighting on sampling results.} As shown in Figure \ref{fig: recon}, it can be observed that the scores of each point, obtained from two different reconstruction strategies, indicate their different points of focus. Therefore, applying different weights to the two scores will directly impact the sampling results. As shown in Figure \ref{fig: vis3}, we tested the impact on the sampling results when different weights are applied, where $\alpha$ represents the weight of $Score_{point}$ and $1-\alpha$ represents the weight of $Score_{shape}$. It can be seen from the figure that there are obvious differences in the sampling results of the two strategies: Point reconstruction focuses more on the contours of objects. In contrast, shape reconstruction pays more attention to flat regions. By applying different weights we obtain rich and diverse sampling results.

\section{Conclusion}
We introduce REPS, an uncomplicated and efficient approach for downsampling point clouds. Through reconstruction at both the point and local shape levels, we can accurately assess the importance of each point using reconstruction loss, thus completing the sampling process effectively. Experimental results confirm the superior downsampling performance of our method for point clouds, demonstrating its adaptability to diverse task requirements. Furthermore, we introduce GLFA, which seamlessly integrates local and global structural features, facilitating efficient reconstruction operations and ensuring the high quality of the sampled point cloud. For future research, considering that our reconstruction has solely relied on the spatial coordinates of the point cloud, investigating feature reconstruction for each point could be beneficial. Additionally, integrating coordinate reconstruction with feature reconstruction could lead to improved sampling outcomes. 

\clearpage
%
%
\bibliographystyle{splncs04}
\bibliography{egbib}
\end{document}


\pagestyle{headings}
\mainmatter
\def\Number{1516}  

\title{Appendices}

\titlerunning{Meta-Sampler} 
\authorrunning{Meta-Sampler} 
\author{Ta-Ying Cheng, Qingyong Hu, Qian Xie, Niki Trigoni, Andrew Markham}
\institute{Department of Computer Science, University of Oxford}

\maketitle

\section{Limitations}
While this work shows the effectiveness of incorporating multiple tasks in the meta-training step, the question of finding the best combination and the number of tasks for the best model is indeed intriguing and important. Intuitively, more tasks will supposedly lead to more exposure to different task varieties. Simultaneously, this could also lead to tasks potentially contradicting each other and thus diminishing the effectiveness of the pretrained meta-sampler.

\begin{figure}[ht]
	\begin{center}
		\includegraphics[width=0.95\linewidth]{Images/vis_1.png}
	\end{center}
	\vspace{-0.1cm}
    \caption{\textbf{Visualisations of sampled points for classification (red), reconstruction (blue), and shape retrieval (green) at sampling ratio of 32.}}
	\label{fig: vis}
\end{figure}

\begin{figure}[t]
	\begin{center}
		\includegraphics[width=0.8\linewidth]{Images/vis_2.png}
	\end{center}
	\vspace{-0.1cm}
    \caption{\textbf{Visualisations for SampleNet trained with single-task single-model (blue) and with single task multi-model (red) at sampling ratio of 32.}}
	\label{fig: vis2}
\end{figure}


Another interesting direction is the sampling strategy for tasks such as large-scale segmentation, where the conventional sampling approaches are progressively used in the network, instead of simply using it once at the beginning. How to adapt our meta-sampler to such tasks efficiently and effectively is worth exploring.

\section{Qualitative Results}

\subsection{Sampling for Different Tasks}
We provide additional qualitative analyses on the different points sampled from different tasks for classification, reconstruction, and shape retrieval (Figure \ref{fig: vis}). The results further validate our assumption that different tasks have different preferences in terms of feature sampling (\textit{e.g.,} classification often require the sampled points to be scattered to give an overview of the object; reconstruction focuses on denser points to minimise the Chamfer Distance loss during reconstruction; shape retrieval often preserve fine-grained features like corners to distinguish hard negative of the same class), and hence an almost-universal sampler is indeed the best way to tackle the task of sampling. 

\subsection{Sampling from Single and Multi-model training}
We visualise the points sampled from single-task single-model (blue) and our proposed multi-task-multi-model training (red), as shown in Figure \ref{fig: vis2}. It is apparent that while there exists some overlapping points (which is intuitive as they are targeting the same task), the majority of points are different. Based on the accuracy improvements from our main paper, we can thus infer that our joint-training scheme allows us to capture features that are more universal towards a particular task.

\section{Reproducibility}
All our training (both meta-sampler and task adaptation) has a batch size of 24. Meta-training uses 5 gradient steps in the inner update. The rapid task adaptation uses the Adam optimiser with a learning rate of 1e-3. The ratio of task, simplification, and projection losses for task adaption is 1:1:1.

We provide the code for our meta-sampler trained on the three tasks, as well as the code for single-task multi-model training on SampleNet in our supplementary. Checkpoints and code will be made publicly available upon the publication of the work.